# Boosted Multiple Kernel Learning for First-Person Activity Recognition

Fatih Özkan[1], Mehmet Ali Arabacı[1], Elif Surer[1], Alptekin Temizel[1,2]
{fatih.ozkan, mehmet.arabaci, elifs, atemizel}@metu.edu.tr

[1]Graduate School of Informatics  
Middle East Technical University  
06800, Ankara, Turkey

[2]Electronic, Electrical and Systems Engineering  
University of Birmingham  
Edgbaston, Birmingham, B15 2TT, UK

*Abstract*—Activity recognition from first-person (ego-centric) videos has recently gained attention due to the increasing ubiquity of the wearable cameras. There has been a surge of efforts adapting existing feature descriptors and designing new descriptors for the first-person videos. An effective activity recognition system requires selection and use of complementary features and appropriate kernels for each feature. In this study, we propose a data-driven framework for first-person activity recognition which effectively selects and combines features and their respective kernels during the training. Our experimental results show that use of Multiple Kernel Learning (MKL) and Boosted MKL in first-person activity recognition problem exhibits improved results in comparison to the state-of-the-art. In addition, these techniques enable the expansion of the framework with new features in an efficient and convenient way.

*Keywords—multiple kernel learning; kernel boosting; first-person; ego-centric videos; activity recognition*

## I. INTRODUCTION

Activity recognition is considered as a supervised learning problem which consists of an activity representation model and an activity classification method. The activity class in the query video is determined based on a dictionary of labelled activity samples. The performance of an activity recognition system is mainly dependent on the effectiveness of the representation model and the accuracy of the classification method.

Third-person videos are captured using a camera pointing at the person or a group of persons. The camera itself is not involved in the action and it is usually static, as opposed to the first-person videos. In the first-person perspective, the camera is attached to the actor - usually to the head or to the chest of the person. The actor is involved in the events and the camera undergoes large amounts of ego-motion such as moving up or down and turning with the activity of the user. This requires taking the ego-motion and the distinctive perspective of the videos into consideration while tackling first-person activity recognition problems. In the activity representation model, the ego-motion is modelled through the use of global features and the other motion occurring in the field of view of the camera are modelled through local features. In [1], Histogram of Optical Flow (HOF) is used as a global feature and 3D XYT space-time features are used as local features, then multi-channel kernels in SVM are used in order to combine these features for activity recognition. In [2], dense optical flows, Local Binary Patterns (LBP) are used as global features and cuboid and Spatial-Temporal Interest Points (STIP) are used as local features. These features are used in a similar classification setting to recognize animal (dog) activities. In [3], a number of new multi-dimensional motion-based descriptors are applied. These global descriptors are used together by concatenating them in a vector in SVM and comparable results to [1] and [2] are obtained without using any local features. Recently, first-person vision activity recognition has become a research topic of interest also for robotics applications to identify activities in the field of view of the robot [4]. In these applications, different from the traditional works, the data can be acquired using RGB-D sensors as the applications are restricted to indoor applications where use of RGB-D sensors are feasible. Availability of depth information allows the use of depth-based descriptors and skeletal data. In the classification stage, Radial Basis Function (RBF) kernels are used in SVM and the kernel parameters are set by cross-validation.

Use of MKL in a multimodal setting to fuse different audio and video features has been proposed for event detection in web videos [5]. It has been shown that MKL performs well even when redundant features are used and it outperforms other popular methods such as wrappers, filters and boosting as well [6]. MKL has been shown to produce promising results during the identification of emergent leaders in meeting scenarios [7].

In the literature, the methods proposed for first-person activity recognition uses a variety of features having distinctive information, which needs to be combined. Use of a standard learning framework without particular extension on feature selection and weighting implies the use of pre-set rules, such as unweighted sum, which gives equal preference to each feature independent of its classification ability. In [1], use of multi-channel kernels is proposed to combine global and local features. Each feature is considered as a separate channel and a pre-defined rule using exponents is utilized to combine them. Use of multimodal Fisher kernel vectors has been investigated to combine video and sensor features for ego-centric activity recognition [8]. In [3], all the features are concatenated in a single vector which is then fed into the learning algorithm, essentially giving each feature equal weight. The detailed analysis of adding and removing features demonstrates their effects on the performance, but the optimal combination of the

This work is supported by The Scientific and Technological Research Council of Turkey under TUBITAK BIDEB-2219 grant no 1059B191500048.

features is left as an open research question. While different types of kernels are evaluated, use of different kernel parameters is not investigated and these kernel variations are not combined. It can be argued that the performance could be improved by using a combination of different kernels and kernel parameters.

This study proposes an innovative approach for first-person activity recognition using MKL and Boosted MKL methods. The novelty of this study lies in its dynamic and adaptive structure which fuses different features and different kernels in an optimized way instead of using pre-determined weights. Dynamic weighting of complementary features is managed through a data-driven approach at the training stage. This solution allows to create an adaptive framework which could easily be customized for specific first-person activity recognition problems as it allows seamless integration of the features and expansion with new features.

The structure of the paper is as follows: in Section 2, global and local motion-related features are explained. MKL and Boosted MKL are described thoroughly in section 3. Experimental results are presented in section 4 and section 5 concludes the study.

## II. GLOBAL AND LOCAL MOTION-RELATED FEATURES

In this section, global and local motion-related features, extracted from the first-person videos, are described. Global descriptors capture basic motion information such as the ego-motion and local descriptors provide the complementary local information, which is necessary for the recognition of different types of activities. In this study, global motion is represented using two descriptors both based on optical flow information: Histogram of Optical Flow (HOF) [1] and Log Covariance (Log-C) [9]. Local information is represented using Cuboids [10].

### A. Histogram of Optical Flow (HOF)

After calculation of the optical flow, the histogram of optical flows in dense grid of $s$ by $s$ cells, is computed. For each cell, optical flows are accumulated with respect to their orientations into 8 representative direction bins to create a histogram. Thus, we obtain $s$ x $s$ x 8 sized histogram bins from each video representing both spatial and temporal information.

### B. Log-Covariance (Log-C)

Log-covariance (Log-C) descriptor is originally designed for third-person videos to capture different characteristics of the motion [10]. For this purpose, at each pixel, 12 dimensional optical flow-based motion-related features and intensity-based gradient vectors are extracted. These are calculated by intensity gradient of raw video sequences with respect to temporal $t$ direction and first-order partial derivative of optical flow with respect to spatial $x$ and $y$ directions, spatial divergence, vorticity, gradient tensor and rate of strain tensor. This set of spatio-temporal features represents dynamics of the motion in first-person videos in a more comprehensive way than basic optical flow-based features. Then, compact covariance descriptors are created by capturing these features in the covariance matrix [9] since high dimensional feature vectors are not efficient for clustering and classification operations.

We use matrix logarithm [11] operation to convert manifold of covariance matrices into Euclidean space since covariance matrices lie on the Riemannian manifold.

### C. Cuboids

Cuboid features are sparse 3D XYT space-time features [10] and they have been used extensively for recognizing behaviour in third-person camera perspectives. Sparse space-time features have been shown to perform well for activity recognition applications [12].

First, spatio-temporal Cuboid feature detector is run in order to detect feature locations. While the idea is similar to spatial detectors, detection proceeds along the temporal direction $t$ in addition to the spatial $x$ and $y$ directions. Then, at each interest point, spatio-temporally windowed pixel values (i.e. flattened gradient vectors) are calculated to form a Cuboid. The Cuboids are specifically designed for behaviour recognition applications and they aim to detect too many features rather than too few in order to handle challenging conditions.

### D. Feature Clustering

The motion information of a video by word occurrences is described by using the bag of visual words (BoW) approach. Each collection of descriptors is separately clustered into multiple types by K-Means algorithm. Thus, each descriptor is assigned to a visual word and the histograms for each video are computed so that representative visual word histograms of each video are obtained. Since each set of feature descriptors is clustered separately, three histograms are computed for each video. The histogram $H_{id}$ is a $w$ dimensional vector for the $i^{th}$ video obtained using descriptor $d$ and $w$ is the number of visual words. For each video, each descriptor histogram computed is concatenated and final histogram is obtained. $H_i=[h_{i1}, h_{i2}, h_{i3}, …, h_{iw}]$ is the histogram of video $v_i$, $h_{iw}$ is the number of $w^{th}$ visual word of the $i^{th}$ video.

## III. MULTIPLE KERNEL LEARNING

In this study, we use both Multiple Kernel Learning (MKL) and Boosted MKL, whose details are explained in this section.

### A. Multiple Kernel Learning

General practice in vision applications presumes a pre-defined parametric kernel and the parameters of the kernel function is determined by cross-validation. Instead of using a single kernel or combining a number of kernels having pre-defined rules as in multi-channel kernels, MKL aims to fuse different features and kernels in an optimal setting. Traditional MKL methods use a convex combination of kernels where all coefficients are non-negative and sum to 1. In parallel to the training of the model, weights of each kernel are optimized and these weights are then used in the final classifier. Effectively, MKL provides a data-driven solution for feature selection and weighting.

Given that $M$ represents features and $L$ is the number of training data in a 2-class classification setting, we have:

$$\{(x_i, y_i)\}_{i=1}^{L}, x_i = \{(x_{i,1}, x_{i,2}, …, x_{i,M})\}, y_i \in \{1, -1\}, \quad (1)$$

where $x_{i,m}$ are the feature vectors for feature $m = \{1,2,...,M\}$ and $y_i$ are the class labels. It has to be noted that the feature vectors for different features might have different dimensions.

For each feature $m$, pairwise differences are measured by a kernel function $K_m(x_i, x_j)$ and as a result, we have a set of $M$ kernels: $\{K_m\}_{m=1}^{M}$.

**Algorithm 1: Boosted MKL**

Inputs: $(x_1, y_1), ..., (x_L, y_L), Training\ data$;
$K_m(x_t, x_n), Kernel\ function$; $c_n, n^{th}\ classifier$;
$C, No.of\ classifiers$; $M, No.of\ kernels$; $T, Trial\ number$;
$initial\ set\ of\ probabilities\ P_1(i) = \frac{1}{L}, i = 1 ... L$
Output: Kernel weight vector $w_{t,n}$ $n^{th}$ classifier of trial $t$;
Prediction labels for the videos computed based on
$$sign(\sum_{t=1}^{T} w_{t,n=1:K} c_{t,n=1:C})$$
$for\ t = 1, ..., T\ do$
  $Select\ n\ videos\ based\ on\ set\ of\ probabilities\ P_t$
  $for\ m = 1, ..., M\ do$
    $Train\ each\ weak\ classifier\ using\ K_m$
    $Compute\ the\ error\ based\ on\ P_t$:
$$e_t = \sum_{i=1}^{L} P_t(i)(c_{t,m}(x_i) \neq y_i)$$
  $end\ for$
$Select\ the\ classifier\ that\ gives\ the\ minimum\ error\ e_t$
$between\ errors\ of\ all\ trials$:
  $e_t = min\ e_{t,m}$
  $Update\ weights\ of\ classifiers$:
$w_t = ln\frac{1-w_t}{w_t}$
  $Update\ P_{t+1}(i)$:
$$P_{t+1}(i) = P_t(i) \times \begin{cases} e^{-w_t}, & if\ c_t(x_i) = y_i \\ e^{w_t}, & if\ c_t(x_i) \neq y_i \end{cases}$$
$end\ for$

Learning is formulated as optimizing the coefficients $\{\alpha_i\}_{i=1}^{L}$ and $b$:

$$f(x) = \sum_{i=1}^{L} \alpha_i y_i K(x, x_j) + b \quad (2)$$

Then, $L \times L$ kernel matrix of pairwise comparisons $k_{i,j}(x_i, x_j)$ is constructed from the data. All kernel methods are designed to process such square matrices.

MKL is formulated as optimizing the kernel weights $\{c_m\}_{i=1}^{M}$:

$$K(x_i, x_j) = \sum_{m=1}^{M} c_m K_m(x_{i,m}, x_{j,m}) \text{ where } c_m \geq 0, \quad (3)$$
$$\sum_{m=1}^{M} c_m = 1$$

It has to be noted that in addition to using different features together, a number of different kernels for the same feature could also be used in this framework.

Since the descriptors mentioned in section II represent different types of motion information, robust combination of these descriptors gains utter importance. To do so, instead of giving the same weight to each feature in classification, MKL could be used in order to assign a self-optimized kernel-feature combination for the classification problem in hand. SimpleMKL simplifies the optimization process based on mixed-norm regularization [13]. It uses gradient descent and iteratively determines the combination of kernels based on a standard SVM solver.

*B. Boosted Multiple Kernel Learning*

Boosted Multiple Kernel Learning (Boosted MKL) is an iterative approach to combine features and kernels effectively. This method aims to integrate the traditional AdaBoost iterative technique via a classifier learning with boosting trials and it has been shown to perform well in a variety of problems [14], [15].

The details of the boosting process used in this paper are given in Algorithm 1. First, classifiers are trained with features extracted through boosting trials. At each trial, training samples are selected from all training videos based on a probability distribution and after each trial, weights of misclassified samples (videos) are increased. Best classifiers of each trial are then used to train all the training samples. Their trial performance is based on their weights.

MKL and boosted MKL have the following advantages:
• Different kernels could be used for different features
• Kernels having different parameters (such as the degree of the polynomial and variance of Gaussian) could be used together
• Each kernel is assigned a weight which gives different weights to different features and kernels

Since a variety of features having different data ranges is used, the feature vectors need to be first normalized.

IV. EXPERIMENTAL RESULTS

In this section, outcome measurements of the experiments and their discussion are presented in detail.

*A. Outcome Measurements*

We conduct experiments in order to evaluate the Boosted MKL, SimpleMKL and traditional multi-channel kernel approach described in [1] and [2] in terms of classification accuracy. We compared these three approaches on the segmented videos of JPL-Interaction [1] and DogCentric activity [2] datasets. There are 84 videos and 7 unique activities available in JPL dataset. At each iteration, we randomly selected 9 training and 3 testing videos for each unique activity so that the dataset is split into 63 training and 21 testing videos. In DogCentric dataset, there are 209 videos and 10 unique activities. At each iteration, we randomly select half of the videos of each activity for training and the rest of the videos for testing. Each experiment has been repeated for 100 times and average classification accuracies at the end of all iterations are reported. Fig. 1 shows the confusion matrices of the activity recognition results using HOF, Log-C and Cuboid features and classification results of SimpleMKL and Boosted MKL based on activities.

In addition to Gaussian and Histogram Intersection (H-Int) [16] kernels, a modified histogram intersection kernel (DC-Int) [2] was used for both datasets. In addition (JPL-Int) [1] kernel is also used for the JPL dataset.

Average classification accuracies of the approaches applied on JPL dataset are shown in Table I. Table II shows the average classification performances of the approaches applied on DogCentric dataset. Table III shows a summary of accuracy results both for two and three-feature sets of SimpleMKL and Boosted MKL, including their comparison to the already existing methods of the literature on JPL and DogCentric datasets. Fig. 1 illustrates the confusion matrices using a single feature and multiple feature combinations.

*B. Discussion of the Results*

As shown in Table I and II, when only a single feature is used, Log-C outperforms HOF and Cuboid for all kernel types. Combining different types of features increases the performance and the best performance is achieved when all features are used (except for the Gaussian kernel case).

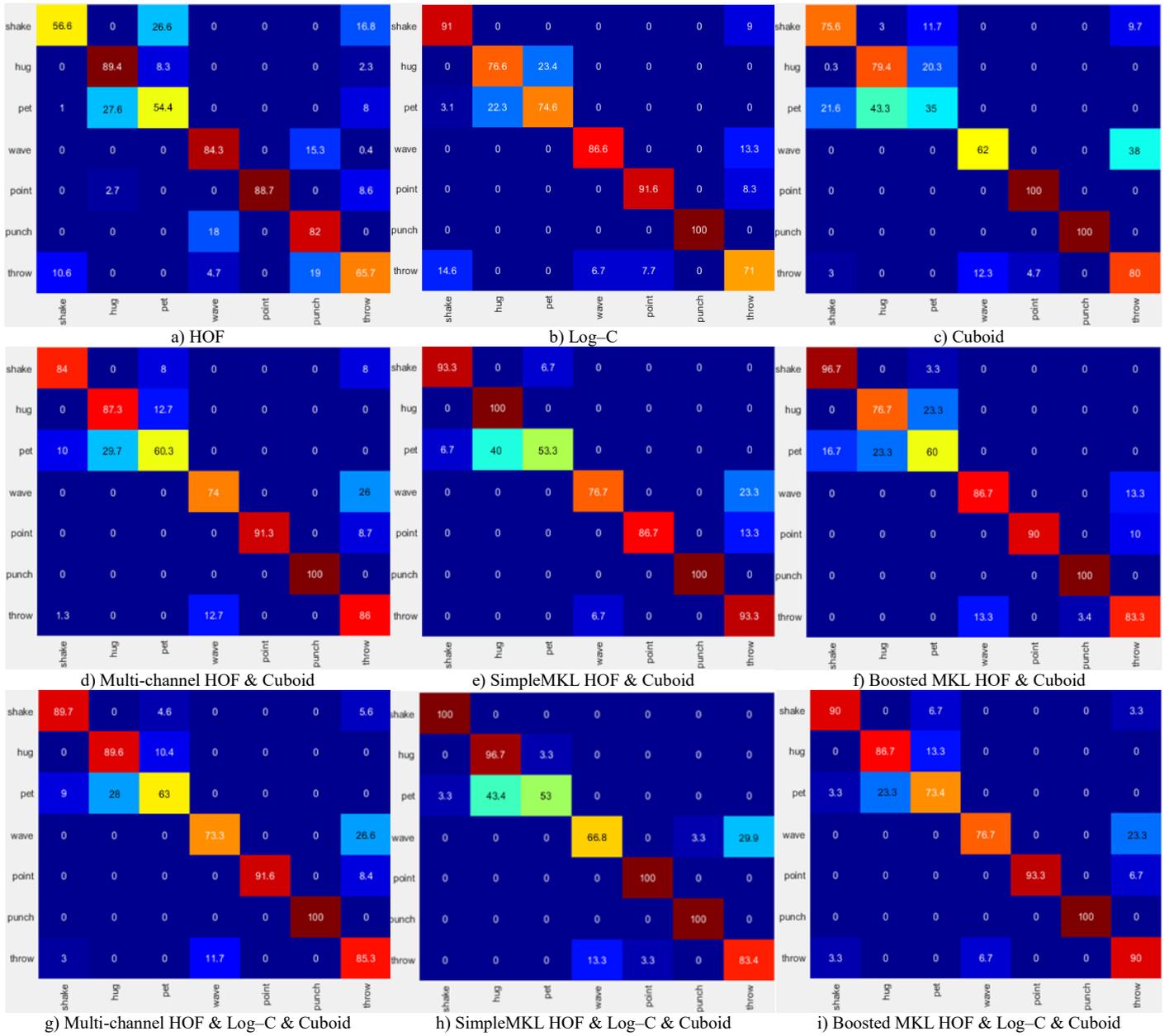

Fig. 1 The confusion matrices of the base and combined features using DC-Int kernel on JPL dataset, SimpleMKL and Boosted MKL.

TABLE I CLASSIFICATION ACCURACIES FOR JPL DATASET

| Features | Kernel Types (%) | | | |
|---|---|---|---|---|
| | *JPL-Int* | *DC-Int* | *H-Int* | *Gaussian* |
| HOF | 70.7 | **74.4** | 66 | 71.8 |
| Log-C | 72.9 | **84.5** | 75.4 | 74.3 |
| Cuboid | 70.8 | **75.6** | 73.6 | 65.0 |
| HOF & Cuboid | **84.2** | 83.2 | 78.2 | 73.8 |
| HOF&Log-C&Cuboid | 82.7 | **84.6** | 78.9 | 75 |

TABLE II CLASSIFICATION ACCURACIES FOR DOGCENTRIC DATASET

| Features | Kernel Types (%) | | |
|---|---|---|---|
| | *DC-Int* | *H-Int* | *Gaussian* |
| HOF | 42.0 | **43.6** | 29.2 |
| Log-C | 43.1 | **46.5** | 29.9 |
| Cuboid | 56.8 | **57.6** | 55.5 |
| HOF & Cuboid | 59.6 | **60.8** | 58.7 |
| HOF&Log-C&Cuboid | 62.2 | **62.3** | 56.6 |

TABLE III ACCURACY RESULTS ON JPL AND DOGCENTRIC DATASETS

| Approaches | Accuracy (%) | |
|---|---|---|
| | DogC dataset | JPL dataset |
| Ryoo et al. [15] | 60.5 | 84.4 |
| Abebe et al. (RMF features) [3] | 61.0 | 86.0 |
| SimpleMKL (2 Features) | **64.9** | 86.1 |
| SimpleMKL (3 Features) | 64.8 | 85.7 |
| Boosted MKL (2 Features) | 62.9 | 82.7 |
| Boosted MKL (3 Features) | 64.2 | **87.4** |

Confusion matrices in Fig. 1a to Fig. 1c demonstrate that different features complement each other since each one is useful for distinct set of activities. When analysed individually, they are good (have higher than or equal to 80% accuracy) for recognition of the following activities respectively for HOF, Log-C and Cuboid: {**hug**, wave, point, punch}, {**shake**, wave, point, punch}, {point, punch, **throw**} and the union of all the sets gives: {hug, wave, point, punch, throw, shake}. Each feature has a specific activity (shown in bold letters) for which the other features are not effective and the union covers all the activities except *pet*. On the other hand, they perform poorly (have lower than 80% accuracy) for the following activities, respectively: {shake, pet, throw}, {hug, pet, throw}, {shake, hug, pet, wave} and their intersection gives: {pet}. There is no activity for which all features perform poorly with the exception of the *pet* activity, which gets confused with the *hug* activity. The overall poor performance on the *pet* may stem from its resemblance with the *hug* activity in the videos since the onset of *hug* involve ego-motion (shaking side to side) similar to the *pet*. Such a resemblance is challenging to the classifier and reduces the overall performance of pet activity's recognition rate. Consequently, when two features (HOF and Cuboid) are combined, there is an observed increase in accuracy values (Fig. 1d to Fig. 1f). Results for other two-feature combinations are not provided for brevity. As expected, combining all three features results in the best performing case (Fig. 1g to Fig. 1i).

Table III shows that use of MKL and Boosted MKL to combine features and multiple kernels improves the results on both datasets compared to the conventional methods. SimpleMKL gives the best results when all the methods are restricted to use two features. Boosted MKL performance increases when all features are used. An analysis for the robustness of the feature combination methods could be done in terms of the standard deviation of the accuracy for each individual activity (diagonal elements of the confusion matrix). A lower standard deviation indicates that the method generalizes for a variety of activities and does not favour specific activities. Standard deviations for the methods HOF, Log-C and Cuboid when all features are used individually are 15.15, 10.65 and 22.57, respectively (average 16.12), indicating that Log-C generalizes for a wider range of activities. When all three methods are used, multi-channel kernels, SimpleMKL and Boosted MKL have standard deviations of 12.46, 19.00 and 9.26, respectively. Indicating that the Boosted MKL becomes expert for a variety of activities by selecting effective kernel combinations to represent different activities and is the most robust method. SimpleMKL results in higher standard deviation than average standard deviation of individual features indicating that it inclines to select kernels favouring some activities.

Boosted MKL can exploit three features rather than two features whereas SimpleMKL gives similar results for both feature sets. MKL and Boosted MKL demonstrate that the feature and kernel selection could be handled in a data-driven way at the training stage eliminating the need for a complicated analysis of performance of individual features and kernel types such as provided in Table I and II.

It has to be noted that in all these cases, the performance could be improved by taking temporal relationships in the video into account (such as structure match proposed in [1]) in the post-processing phase. However, as our aim is to analyse the effects of feature and kernel selection and their combination, we leave such extensions as future work.

## V. CONCLUSIONS

We proposed a novel framework based on data-driven selection and weighting of complementary features using MKL and Boosted MKL for first-person activity recognition domain. This framework currently uses HOF, Log-C and Cuboids as features and allows the integration of other video-based features as well as features extracted from other modalities in a dynamic and adaptive way. Data-driven properties of this framework facilitate selection of the complementary features and kernels automatically during training instead of relying on fixed and static weights. In the future, other features such as virtual inertial data, audio features and multi-dimensional motion features could be integrated into the framework.